\pdfoutput=1

\documentclass[11pt]{article}
\usepackage{booktabs} 
\usepackage{graphicx} 
\usepackage{array}
\usepackage{multirow}
\usepackage{amsmath}

\usepackage{float}
\usepackage{xcolor}
\usepackage{colortbl}

\usepackage[final]{acl}

\usepackage{times}
\usepackage{latexsym}
\usepackage{xspace}

\usepackage[T1]{fontenc}

\usepackage[utf8]{inputenc}

\usepackage{microtype}

\usepackage{inconsolata}

\usepackage{graphicx}

\usepackage{xspace}
\newcommand\qa{\textsc{qa}\xspace}
\newcommand\convqa{\textsc{convqa}\xspace}
\newcommand\qrecc{\textsc{qrecc}\xspace}
\newcommand\orconvqa{\textsc{or-convqa}\xspace}
\newcommand\rag{\textsc{rag}\xspace}
\newcommand\ir{\textsc{ir}}\xspace
\newcommand\llm{\textsc{llm}\xspace}
\newcommand\dialogllm{\textsc{dialog-llm}\xspace}
\newcommand\tfive{\textsc{t5}\xspace}
\newcommand\mamba{\textsc{mamba}\xspace}
\newcommand\gpttwo{\textsc{gpt-2}\xspace}
\newcommand\gptfour{\textsc{gpt-4}\xspace}
\newcommand\llamaeightb{\textsc{llama-8b}\xspace}
\newcommand\claude{\textsc{claude-sonnet}\xspace}
\newcommand\mparam{\textsc{m}\xspace}
\newcommand\bert{\textsc{bert}\xspace}
\newcommand\va{\textsc{va}\xspace}
\newcommand\dmv{\textsc{dmv}\xspace}
\newcommand\ssa{\textsc{ssa}\xspace}
\newcommand\doctodial{\textsc{doc2dial}\xspace}
\newcommand\multidoctodial{\textsc{multidoc2dial}\xspace}

\newcommand\rrf{\textsc{rrf}\xspace}
\newcommand\minilm{Mini\textsc{lm}\xspace}
\newcommand\bleu{\textsc{sacrebleu}\xspace}
\newcommand\sbleu{\textsc{sbleu}\xspace}
\newcommand\map{\textsc{map}\xspace}
\newcommand\bm{\textsc{bm}25\xspace}
\newcommand\meteor{\textsc{meteor}\xspace}
\newcommand\bertscore{\textsc{bertscore}\xspace}

\newcommand\pl{\textsc{pl}\xspace}

\newcommand\user{\textsc{user}\xspace}
\newcommand\system{\textsc{system}\xspace}
\newcommand\qrelscore{\textsc{qrelscore}\xspace}
\newcommand\quantidce{\textsc{quantidce}\xspace}

\title{Building Open-Retrieval Conversational Question Answering Systems by Generating Synthetic Data and Decontextualizing User Questions 
}

\author{
 \textbf{Christos Vlachos\textsuperscript{1,2}},
 \textbf{Nikolaos Stylianou\textsuperscript{2}},
 \textbf{Alexandra Fiotaki\textsuperscript{2}},
 \textbf{Spiros Methenitis\textsuperscript{2}},
\\
 \textbf{Elisavet Palogiannidi\textsuperscript{4}},
 \textbf{Themos Stafylakis\textsuperscript{1,2,3}},
 \textbf{Ion Androutsopoulos\textsuperscript{1,3}}
\\
\\
 \textsuperscript{1}Department of Informatics, Athens University of Economics and Business, Greece,\\
 \textsuperscript{2}Omilia - Conversational Intelligence,
 \textsuperscript{3}Archimedes, Athena Research Center, Greece,\\
 \textsuperscript{4}NCSR Demokritos, Greece
\\
 \small{
   \textbf{}
 }
}

\begin{document}
\maketitle
\begin{abstract}
We consider open-retrieval conversational question answering (\orconvqa), an extension of question answering where system responses need to be (i) aware of dialog history and (ii) grounded in documents (or document fragments) retrieved per question.
Domain-specific training datasets for \orconvqa are crucial for real-world applications but difficult to obtain. We propose a pipeline that capitalizes on the abundance of plain text documents in organizations (e.g., product documentation) to automatically produce realistic \orconvqa dialogs with annotations. Similarly to real-world human-annotated \orconvqa datasets, we generate in-dialog question-answer pairs, self-contained (decontextualized, e.g., without referring expressions) versions of user questions, as well as propositions (sentences expressing prominent information from the documents) in which the system responses are grounded. We show how synthetic dialogs can be used to train efficient question rewriters or retrievers. The retrieved information can then be passed on to an \llm that generates the system response.
\end{abstract}

\section{Introduction} \label{sec:introduction}
Retrieval-Augmented Generation (\rag) is used to ground large language models (\llm{s}) to knowledge outside of their training data and limit hallucinations \cite{Lewis-etal-2020-rag}. \rag is especially applicable to conversational agents, enabling them to provide responses grounded in retrieved documents. We focus on open-retrieval conversational question answering (\orconvqa), an extension of question answering where system responses need to be (i) aware of the dialog history and (ii) grounded in the retrieved documents (retrieved per question). 

Compared to conventional Information Retrieval (\ir) \cite{Manning_Raghavan_Schütze_2008_IR}, \orconvqa introduces two challenges. Firstly, the system needs to account for the additional context of the dialog \cite{Mao-etal-2022-curriculum}, mostly the dialog history (previous system and user turns). Solely relying on the last user question to query the document repository may result in suboptimal answers, since discourse phenomena like ellipsis and co-reference are prevalent in dialogs \cite{jurafsky-etal-2024-book, Dalton-etal-2022-conv-info-seeking, Zaib-etal-2022-qa_survey, Zamani-etal-2022-cis}. Thus, the dialog history has to be considered jointly with the last user question, which becomes an issue when the history includes information irrelevant to the last question, or the history is too long. Alternatively, a separate model may produce a self-contained (`decontextualized') version of the last user question (e.g., with no ellipsis, anaphora) \cite{li-gaussier-2024-domain, Yu-etal-2020-few-rewriting, chieh-etal-2020-conv-rewriting}, allowing the use of existing dialog-unaware retrievers, which expect a stand-alone query. This approach, \emph{query reformulation}, either rewrites the user question to include all relevant information 
or appends relevant tokens 
from the dialog history \cite{mo-etal-2023-convgqr}.

A second challenge in \orconvqa is the lack of domain-specific data and annotations \cite{mo-etal-2024-convsdg}, which are crucial to train real-life systems. 
Collecting and especially manually annotating new dialog data for specific domains is particularly cumbersome. Alternatives, such as Dialog Inpainting \cite{dai-etal-2022-inpainting, hwang-etal-2023-dialogizer, wu-etal-2024-synthesizing} or synthesizing dialogs from scratch \cite{kim-etal-2022-generating}, generate synthetic data 
from domain-specific documents, which are abundant in practice (e.g., product documentation, recommendation guidelines). However, previous alternatives of this kind make unrealistic assumptions, such as presuming a one-to-one correspondence between document sentences and possible user questions, and/or assuming that additional manually annotated dialogs are available to fine-tune system components \cite{dai-etal-2022-inpainting,hwang-etal-2023-dialogizer}.

Motivated by such issues, we propose a pipeline to generate realistic synthetic, document-grounded \orconvqa dialogs and annotations. Like previous approaches, we use domain-specific documents, but without requiring \emph{any} additional training data and without assuming a one-to-one mapping between document sentences and user questions.

The pipeline first prompts an \llm, to generate \emph{propositions} from the documents in the repository. Similarly to \citet{chen-etal-2024-dense}, we require each proposition to be a stand-alone simple sentence (e.g., without compound sentences, anaphora or ellipsis) expressing information from a document. Unlike \citet{chen-etal-2024-dense}, however, we require each proposition to convey information important enough to be requested by a user query. Some document sentences may not be used in any of our propositions (hence, they may not be used to answer any question), and some questions may require information from multiple propositions. The retrieval pool may then contain the propositions, not the original documents or document fragments, making it easier to retrieve the information needed by a user question, without irrelevant information. 

The pipeline then prompts the same \llm to generate \orconvqa dialogs from sampled propositions. Each dialogic pair (user and system turn) includes a contextualized (dependent on dialog history) and decontextualized (self-contained) version of the user question, the corresponding system response (answer), and the propositions used to generate the question and the response.

We experimentally show the superiority of dialogs generated through our propositions compared to using 
directly document sentences, by measuring the coherence of the dialogs, their relevance to the knowledge they are grounded in, and improvements in retrieval scores. To demonstrate the usefulness of the generated synthetic dialogs, we show how they can be used to fine-tune light models as question rewriters or retrievers. 
The retrieved 
information and the user questions are then given 
to an \llm 
that produces the system response. We verify the effectiveness and efficiency of our fine-tuned models on both synthetic and real-world test data,
comparing against rewriting questions by prompting larger \llm{s},
or using the last user query (with or without concatenating the dialog history). We also propose a new mechanism to detect questions that are already self-contained and do not require rewriting, improving inference speed further. We leave for future work the question of how to use synthetic data to fine-tune lighter response generation models too, instead of prompting larger \llm{s} for response generation.

Overall, our main contributions are: (1) We propose a pipeline to create high-quality synthetic annotated \orconvqa dialogs from domain-specific documents, without requiring \emph{any} manually annotated training data and without assuming a one-to-one mapping between document sentences and user questions. (2) We demonstrate the superiority of synthetic dialogs generated by first converting the documents to propositions that capture important information, compared to directly using document sentences. (3) We show how the generated synthetic dialogs can be used to fine-tune light question rewriters or retrievers. (4) We make publicly available our source code and a synthetic \orconvqa dataset to facilitate future research.\footnote{\url{https://github.com/christosvhs/Document-Grounded-Conversational-QA}}

\section{Related Work} 
\subsection{Conversational Question Answering}
In the simplest case, Conversational Question Answering (\convqa) systems answer a \emph{sequence} of questions about a \emph{single} given (always the same) document, by identifying spans of the document that answer each question. The difference from machine-reading comprehension datasets such as \textsc{squad} \cite{rajpurkar-etal-2016-squad} is that the context includes not only the document, but also the previous questions and answers. \citet{choi-etal-2018-quac} and \citet{reddy-etal-2019-coqa} concatenate the document with the last $k$ dialog turns, and fine-tune an encoder to predict the document span that answers the last user question. 
In similar work, \citet{huang-etal-2018-flow}, \citet{yeh-chen-2019-flow2}, \citet{Zhu-etal-2018-SDNet}, \citet{qu-etal-2019-hae}, \citet{campos-etal-2020-doqa} 
use additional intermediate representations 
of the encoder.  


In \orconvqa, the system again needs to take into account the dialog history, but it also needs to retrieve relevant documents for each user question and compose an answer, typically by feeding the retrieved information to an \llm. For retrieval, one may again concatenate the last $k$ dialog turns to obtain queries that include the dialog history, and fine-tune a retriever to handle queries of this kind \cite{qu-etal-2020-orquaq, anantha-etal-2021-qrecc}. 
Fine-tuning the retriever, however, typically requires  training data with ground truth, which are difficult to obtain. Thus, zero-shot 
\cite{krasakis-etal-2022-zero-rag} and approaches with limited supervision \cite{qu-etal-2020-orquaq-limited, Voskarides-etal-2020-query-resolution, li-gaussier-2024-domain, Mao-etal-2022-curriculum} have also been proposed.      


\subsection{Query reformulation} \label{sec:reformulation}

Instead of fine-tuning the retriever to handle queries that include the dialog history, it is computationally cheaper and requires less data \cite{wu-etal-2022-conqrr, zhang-etal-2024-adaptive} to train a question rewriter to decontextualize (i.e., to make self-contained) the user questions. This allows utilizing existing dialog-unaware retrievers, which expect stand-alone questions as queries, without fine-tuning them.  

Question rewriting is the dominant approach to handle dialog history in \orconvqa and, more generally, \convqa, to the point that it is treated as a task of its own \cite{elgohary-etal-2019-unpack}. Most question rewriting  approaches leverage Transformers \cite{vaswani-etal-2017-attention} fine-tuned on datasets such as those of \citet{anantha-etal-2021-qrecc}, \citet{elgohary-etal-2019-unpack}, \citet{ren-etal-2021-saac}, which include user questions and ground truth rewrites \cite{li-gaussier-2024-domain, Yu-etal-2020-few-rewriting, chieh-etal-2020-conv-rewriting, Vakulenko-etal-2020-Question-Rewriting}. \citet{cheng-etal-2024-interpreting} propose a multitask approach for both retrieval and query rewriting (rewrite the query so that it includes all relevant information from the dialog history). \citet{mo-etal-2023-convgqr} perform both question rewriting and query expansion (adding history tokens). \citet{Mo-etal-2023-learning} train their model to identify dialog turns complementary to the last user question.

Query reformulation can also be achieved implicitly. \citet{Yu-etal-2021-implicit-contectualization} use \bert \cite{devlin-etal-2019-bert} to encode the last user question concatenated with the dialog history. They also encode the ground truth query reformulation using the query encoder of an ad-hoc retriever. They fine-tune \bert to minimize the mean squared error loss of the two encodings, in addition to the ranking loss of the \bert encoding of the user question and dialog history. Reinforcement learning has also been leveraged for question rewriting \cite{wu-etal-2022-conqrr, ma-etal-2023-query}. Finally, rewrites can also be generated via prompting \llm{s} using few or no examples \cite{mao-etal-2023-llm-rewrite1, ye-etal-2023-llm-rewrite2, yoon-etal-2024-llm-rewrite3}.

\subsection{Synthetic data generation for ConvQA}
There is a plethora of manually annotated \convqa datasets \cite{elgohary-etal-2019-unpack, anantha-etal-2021-qrecc, choi-etal-2018-quac, qu-etal-2020-orquaq, ren-etal-2021-saac, reddy-etal-2019-coqa, campos-etal-2020-doqa, feng-etal-2020-doc2dial, feng-etal-2021-multidoc2dial}, but such volumes of annotated data are expensive to compile and scarce in practice when moving to new application domains. 

A promising direction to alleviate this issue in \orconvqa is to leverage domain-specific documents. In Dialog Inpainting, consecutive sentences of a document are considered an answer to a user question that an \llm  tries to generate \cite{dai-etal-2022-inpainting, hwang-etal-2023-dialogizer, wu-etal-2024-synthesizing}. Contrary to our work, this approach assumes that every sequence of sentences is an answer to a possible user question; in practice, however, some document parts may not convey information users would be interested in. In the original Dialog Inpainting, a question generation model also needs to be trained, which requires additional annotated data. 

\citet{huang-etal-2023-converser} also generate synthetic questions by prompting an \llm. They feed, however, the \llm with ground truth passages (answering user questions) from existing datasets, which are again difficult to obtain in new application domains. They also consider only retrieval, not response generation. 
\citet{mao-etal-2022-convtrans} generate dialog questions from existing web searches. \citet{mo-etal-2024-convsdg} instruct an \llm to generate dialogs around certain topics, which results in dialogs not grounded in specific documents. In similar work, \citet{bitton-etal-2023-q2d} utilize user questions from publicly available \qa datasets, instead of topic descriptions. 

Closer to our pipeline is the work of \citet{kim-etal-2022-generating} and \citet{liu-etal-2024-chatqa}. The former identifies document fragments that may provide answers to possible user questions, from which synthetic questions and answers are extracted.  Contrary to our work, however, their pipeline requires additional annotated data to train their question-answer extractors. \citet{liu-etal-2024-chatqa}  provide a \emph{single} document to an \llm and instruct it to generate a dialog. By contrast, our synthetic dialogs can be grounded on propositions from multiple documents.

\section{Methodology}
\subsection{Domain-specific documents} \label{sec:documents}
\label{sec:bootstraping}
We hypothesize that our pipeline will be especially beneficial in scenarios revolving around domain-specific, knowledge-intensive documents, as those used in call centers. Hence, we collect 1,036 documents from call centers, henceforth \emph{proprietary documents}. As real-world documents, they may contain unstructured, irrelevant or no information at all, which the \dialogllm will have to account for. The proprietary documents cover four domains: software, finance, insurance, miscellaneous (misc).
We also leverage the 488 publicly available documents of \doctodial \cite{feng-etal-2020-doc2dial} and \multidoctodial \cite{feng-etal-2021-multidoc2dial}; both datasets use the same documents, hereafter \doctodial or \emph{public documents}, which are similar to the proprietary ones in quantity, origin, and domains. \doctodial documents originate from government websites covering insurance (\va), student financial support (\textsc{studentaid}), car rental (\dmv), and social security services (\ssa). The \doctodial dataset includes 69,820 dialog turns from 4,470 dialogs, while \multidoctodial includes 61,078 turns from 4,796 dialogs, all grounded in the 488 documents provided. All dialogs were created via crowd-sourcing. The main difference between \doctodial and \multidoctodial is that in the latter, dialogs may be grounded in more than one document. We use \emph{only the test dialogs} of both datasets (1,322 dialogs in total), to measure the performance of our methods on real user questions. 

\begin{figure}[t]
\begin{center}
\includegraphics[scale=0.45]{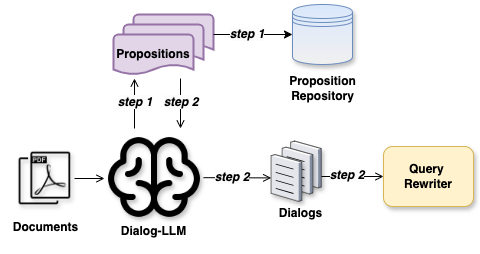} 
\vspace{-3mm}
\caption{Our synthetic dialog generation pipeline.}
\label{figures/pipeline.png}
\vspace{-5mm}
\end{center}
\end{figure}

\subsection{Synthetic dialog generation pipeline}
\label{proposed_pipeline}
Our dialog generation pipeline follows two steps, mainly revolving around prompting LLMs. These prompts were created and refined after rigorous manual evaluation of LLM responses for each part of the process separately.  

\textbf{Step 1}: Following \citet{chen-etal-2024-sub} and \citet{chen-etal-2024-dense}, our synthetic dialog generation pipeline (Fig.~\ref{figures/pipeline.png}) first prompts an \llm , hereafter `Dialog-\llm', to extract propositions from the documents (\S\ref{sec:introduction}).\footnote{The full prompt is provided in Appendix~\ref{sec:appendix_a_step1}.} 
\citet{chen-etal-2024-sub} and \citet{chen-etal-2024-dense} used large commercial \llm{s} (\gptfour) to generate propositions, because smaller and open-source \llm{s} were inadequate for the (complex) proposition generation task. 
We follow their findings and leverage Claude 3.5\footnote{\url{www.anthropic.com/news/claude-3-5-sonnet}} in our experiments, which is a model comparable to \gptfour that \citet{chen-etal-2024-dense} used.

We instruct the Dialog-\llm to split compound sentences into simple ones, separate information into standalone sentences, and decontextualize them to remove references from one proposition to another, taking care to generate propositions only for information that users are likely to ask about, unlike the original propositions of \citet{chen-etal-2024-dense}. 
We obtain 11,566 and 14,443 propositions from proprietary and \doctodial documents, respectively. The propositions of all documents are inserted into a single list, keeping propositions from the same document adjacent. We split the list into non-overlapping sublists of size $n$. Each sublist may, thus, contain propositions from one or more documents. For the proprietary data, we set $n=30$, resulting in each dialog being grounded in 1.5 documents on average, and each question being grounded in 2 propositions on average. We maintain the same $n$ for the publicly available documents for a fair comparison. We also note that the choice of $n$ is not strict; it can be changed without affecting the validity of the generated dialogs.

\smallskip
\noindent\textbf{Step 2} generates synthetic dialogs and annotations (Fig.~\ref{figures/pipeline.png}) by prompting the same \llm (Dialog-\llm) with three separate prompts (Appendix~\ref{sec:appendix_a_step2}).\footnote{Statistics for the generated dialogs shown in Appendix \ref{sec:appendix_stats}.}

\noindent\textbf{Prompt 2.1 (dialog generation):} This prompt instructs the \llm to generate a user-system dialog grounded in a sampled sublist of $n$ propositions (Step 1). 
Each sublist is used only once, to generate a single dialog. In this step, we instruct the Dialog-\llm to ensure the user questions are decontextualized, i.e., that they include all the necessary information from the dialog context. We find that generating decontextualized questions first and then contextualizing them (using Prompt 2.2) leads to dialogs where more user turns have been decontextualized correctly, instead of the opposite.   

\noindent\textbf{Prompt 2.2 (contextualized questions):} The second prompt of Step 2 instructs the Dialog-\llm to create contextualized versions of the user questions, taking into account the dialog context (e.g., inserting pronouns when entities have been mentioned in the dialog history). An example can be seen in Appendix \ref{sec:appendix_dial}. Hence, there are two versions of each user question, the contextualized and the decontextualized one, along with the system response.


\noindent\textbf{Prompt 2.3 (ground truth propositions):} The third prompt of Step 2 feeds the Dialog-\llm with each sublist of propositions and the corresponding generated dialog, and instructs it to identify (generate again) the propositions each question-answer pair is grounded in. Thereafter, each pair will contain two versions of the user question (contextualized and decontextualized), the system response, and the corresponding propositions. The Dialog-\llm is also instructed to generate an additional token (`accepted' or `not\_accepted') for each pair, signifying whether the pair is indeed grounded in the selected propositions or not.
We remove pairs marked as `not\_accepted' and replace each subsequent user question with its decontextualized version, to avoid referring to a removed pair. This seldom happens. Finally, to avoid any potential mismatch between the propositions generated in Step 1 and the ground truth propositions (generated in this step), we use \bm \cite{Robertson-etal-2009-bm25} to replace each propositions generated in this step with its closest match from Step 1.

\subsection{Building domain-specific systems} \label{sec:bulding}
Having defined our dialog-generation pipeline, we now describe a way to leverage it for \orconvqa applications.
To build an \orconvqa system for a new application domain, we first apply our pipeline to the domain-specific documents the user questions will be answered from. This also converts the documents to propositions, stored in the proposition repository (Fig.~\ref{figures/pipeline.png}). The synthetic data is also used to fine-tune a light query rewriter to decontextualize user questions. Then, in real-life dialogs, each user question is decontextualized and fed to an off-the-shelf (not fine-tuned) retriever to obtain relevant propositions from the proposition repository. 
The retrieved 
information and the user question are then given to an \llm, hereafter Response-\llm, instructed (prompt in  Appendix~\ref{sec:appendix_a_step4}) to generate the system response. In our experiments, we use \llamaeightb \cite{dubey-etal-2024-llama} as the Response-\llm without fine-tuning it.
In additional experiments, we investigate directly fine-tuning a lightweight retriever on our synthetic data, without decontextualizing the user questions, adding the dialog history to the last user question.

\section{Experiments} \label{results}
\subsection{Experimental setup} \label{sec:setup}
We experiment with dense retrieval, sparse retrieval, and  Reciprocal Rank Fusion \cite{Cormack-etal-2009-k} (\rrf). We always feed the Response-\llm with the top 20 retrieved propositions. 
For dense retrieval, we use \minilm \cite{wang-etal-2020-minilm} to embed the propositions of the proposition repository (Fig.~\ref{figures/pipeline.png}) and the user questions. For each query we retrieve the top-20 propositions with the highest cosine similarity. For sparse retrieval, we use \bm.
\rrf fuses the scores of the two other retrievers as follows:
\begin{equation*} 
\textit{score}_i = \frac{1}{\textit{rank}_i^b + k} + \frac{1}{\textit{rank}_i^d + k},
\end{equation*}
where $\textit{score}_i$ refers to the new score assigned by \rrf, $i$ is the index of the propositions regardless of rank, $b$ and $d$ refer to  \bm and dense retrieval, respectively, and $k$ is set to 60 as per usual practice \cite{Cormack-etal-2009-k}. We do not tune $k$ further, nor do we assign weights to the two terms.

For every experiment involving synthetic dialogs, we split them into training and test sets using three different seeds, and report average scores on the test sets. The synthetic training sets are only used to train the models described in \S\ref{sec:rewriters} and tune the hyper-parameters of \bm.\footnote{In \bm, $k_1=0.05$, $b=5$. The best rewriter checkpoint is selected on validation data held out from the training set.}
We also use the original test sets of \doctodial and \multidoctodial, unchanged, and conduct the corresponding experiments only once; the training sets of these datasets are not used, since the question rewriter and retriever are always trained on synthetic data, to demonstrate that our approach requires no manually annotated training data. 

To measure retrieval performance, we compute Mean Average Precision (\map), and Recall at the top-$k$ retrieved items (R@$k$). For response generation, we report \bleu (\sbleu) \cite{post-2018-call} measuring 4-grams, \meteor \cite{banerjee-lavie-2005-meteor}, \bertscore \cite{Zhang-etal-2020-BERTScore}, and the perplexity (\pl) of the Response-\llm. We also report additional experiments, each one considering a single domain, in Appendix~\ref{sec:appendix_b}.

\subsection{Propositions vs.\ sentences}
\label{prop_vs_sent}
We hypothesize that converting the domain-specific documents to propositions leads to synthetic dialogs 
of higher quality, compared to dialogs generated directly from document sentences.
To confirm this, we employ the pipeline of Fig.~\ref{figures/pipeline.png} to generate dialogs with both approaches (propositions, sentences), applying it to the proprietary and public (\doctodial) documents (\S\ref{sec:documents}). 
To generate sentence-based dialogs, we split the documents into sentences and form chunks of 30 consecutive ones, matching the size ($n$) of the proposition sublists used to generate dialogs in Step~1 (\S\ref{proposed_pipeline}).\footnote{We use \textsc{nltk} (\url{www.nltk.org/}) for sentence splitting.} From the proprietary documents, we extract 20,520 sentences, which the pipeline uses to generate dialogs; 36\% of proposition-generated and 33\% of sentence-generated user questions require rewriting. From \doctodial documents, we extract 17,197 sentences; 27\% and 28\% of user questions require rewriting, respectively. 

We compare the quality of proposition-based and sentence-based dialogs by measuring the relevance of the dialogs to the knowledge they are grounded in, dialog coherence, and retrieval performance. 

\smallskip
\noindent\textbf{Relevance}: We employ \qrelscore \cite{wang-etal-2022-qrelscore} to measure the  relevance of each synthetic user question to the corresponding ground-truth propositions (Prompt 2.3) or document chunks, and we then average over user questions.
\qrelscore ranges in $[0, 1]$. It is the harmonic mean of two terms. For the first term, the user question is concatenated with its ground-truth propositions or document chunks, and it is fed to an off-the-shelf \bert. For every layer of \bert, the cosine similarities between each token embedding of the question and each token embedding of the ground truth are calculated and
averaged across all layers. The second term measures the difference between the likelihood of an off-the-shelf \textsc{gpt2} \cite{radford-etal-2019-gpt2} generating the context with, and without conditioning on the corresponding question.

\smallskip 
\noindent\textbf{Coherence}: To measure dialog coherence, we use \quantidce \cite{ye-etal-2021-towards-quantifiable}, which considers the dialogs themselves (not the ground-truth propositions or document chunks). \quantidce employs a \bert model fine-tuned for dialog coherence evaluation on a large dialog corpus. It ranges in $[1, 5]$. 

\smallskip
For relevance (\qrelscore), we consider both contextualized and decontextualized user questions. For dialog coherence (\quantidce), we only consider the contextualized questions, as they better mimic real-world dialogs.
Table~\ref{tab:quality} reports the \qrelscore and \quantidce scores. When using the proprietary documents, proposition-based dialogs are clearly better than sentence-based ones in relevance (\qrelscore). When using public (\doctodial) documents, however, both approaches are on par. In dialog coherence (\quantidce),  sentence-based dialogs are slightly better, both with proprietary and public documents, but the differences are minute (recall that \quantidce ranges in $[1,5]$). 
Overall, we conclude so far (Table \ref{tab:quality}) that proposition-based dialogs are on par with sentence-based dialogs and no clear distinction can be made in terms of quality of generated dialogs. To obtain a clear winner between the two approaches, we turn to retrieval performance, which is also our primary task of interest.

\begin{table}[h]
    \centering
    \resizebox{\columnwidth}{!}{
    \begin{tabular}{llccc}
        \toprule
        \textbf{Docs} &  & \textbf{\qrelscore}$\uparrow$ & \textbf{\qrelscore}$\uparrow$ & \textbf{\quantidce}$\uparrow$ \\
        & & \textbf{(co)} & \textbf{(de)} & \textbf{(co)} \\
        \midrule
        \multirow{2}{1.5em}{PR} & Prop & \textbf{0.36} & \textbf{0.41} & 3.16 \\
        & Sent & 0.25 & 0.27 & \textbf{3.18} \\
        \midrule
        \multirow{2}{1.5em}{PU} & Prop & \textbf{0.33} & \textbf{0.36} & 3.08 \\
        & Sent & \textbf{0.33} & 0.35 & \textbf{3.14} \\

    \end{tabular}
    }
    \caption{\textbf{Relevance} (\qrelscore) and \textbf{Coherence} (\quantidce) results for \textbf{synthetic dialogs} generated through \textbf{propositions} (Prop) or \textbf{sentences} (Sent) using proprietary (PR) and public documents (PU). (co): contextualized questions, (de): decontextualized questions.}
    \vspace{-3mm}
    \label{tab:quality}
\end{table}


\noindent\textbf{Retrieval:} We now compare proposition-based to sentence-based synthetic dialogs by comparing retrieval performance. 
We use \rrf to retrieve either propositions or sentences, and compare three query types: concatenation of the dialog history with the last contextualized user question (Context), contextualized user question alone (Query\textsubscript{co}), decontextualized user question alone (Query\textsubscript{de}). We use the previous question-answer pair only as the dialog history, as it led to the best Context results. Note that the decontextualized user questions used here are the `ground-truth' ones generated by Dialog-\llm (in Step 2). Table~\ref{tab:prop_vs_sent_private} shows that proposition-generated dialogs lead to substantially higher retrieval performance, compared to sentence-generated dialogs, which can be attributed to the clearer and more prominent information propositions express.
We consider the superior retrieval performance of proposition-generated dialogs as an indication of higher-quality synthetic data, since ground truth decontextualized questions should lead to high retrieval scores. Consequently, we use proposition-based synthetic dialogs in subsequent experiments. Table~\ref{tab:prop_vs_sent_private} also shows that concatenating the dialog history with the last user question leads to substantially worse retrieval performance (for off-the-shelf retrievers), probably due to the noise that previous utterances may introduce, as pointed out by \citet{Mao-etal-2022-curriculum}.


\begin{table}[h!]
\begin{center}
\resizebox{\columnwidth}{!}{
\begin{tabular}{llllllll}
\hline
                           \textbf{PR} & \textbf{Query}                    & \textbf{MAP$\uparrow$}  & \textbf{R@5$\uparrow$}  & \textbf{R@10$\uparrow$} & \textbf{R@20}$\uparrow$ \\ \hline
\multirow{3}{1.5em}{Prop } & Context                    & 0.19 & 0.31 & 0.44 & 0.56 \\
                           & Query\textsubscript{co}    & 0.46 & 0.55 & 0.63 & 0.69 \\
                           & Query\textsubscript{de}    & \textbf{0.50} & \textbf{0.60} & \textbf{0.67} & \textbf{0.73} \\ \hline
\multirow{3}{1.5em}{Sent } & Context                    & 0.09 & 0.13 & 0.20 & 0.27 \\
                           & Query\textsubscript{co}    & 0.20 & 0.25 & 0.30 & 0.35 \\
                           & Query\textsubscript{de}    & 0.21 & 0.26 & 0.31 & 0.37 \\
\hline
                          \textbf{PU} & \textbf{Query}                      & \textbf{MAP$\uparrow$}  & \textbf{R@5$\uparrow$}  & \textbf{R@10$\uparrow$} & \textbf{R@20$\uparrow$} \\ \hline
\multirow{3}{1.5em}{Prop } & Context                    & 0.18 & 0.30 & 0.44 & 0.56 \\
                           & Query\textsubscript{co}    & 0.49 & 0.59 & 0.65 & 0.72 \\
                           & Query\textsubscript{de}    & \textbf{0.54} & \textbf{0.63} & \textbf{0.71} & \textbf{0.77} \\ \hline
\multirow{3}{1.5em}{Sent } & Context                    & 0.12 & 0.19 & 0.27 & 0.36 \\
                           & Query\textsubscript{co}    & 0.30 & 0.37 & 0.43 & 0.50 \\
                           & Query\textsubscript{de}    & 0.31 & 0.39 & 0.46 & 0.52 
                           
\end{tabular}
}
\caption{\textbf{\rrf retrieval results} in \textbf{synthetic dialogs} generated through \textbf{propositions} (Prop) \textbf{or sentences} (Sent) using proprietary (PR) and public (\doctodial) documents (PU). Context: concatenated last user question and history, Query\textsubscript{co}: contextualized question only, Query\textsubscript{de}: ground-truth decontextualized question only.}
\vspace{-5mm}
\label{tab:prop_vs_sent_private}
\end{center}
\end{table}

\begin{table}[t]
\begin{center}
\resizebox{\columnwidth}{!}{
\begin{tabular}{llllllll}
\hline
                           & \textbf{Method}                      & \textbf{MAP$\uparrow$}  & \textbf{R@5$\uparrow$}  & \textbf{R@10$\uparrow$} & \textbf{R@20$\uparrow$} \\ \hline
\multirow{3}{1.5em}{PR}   
                           & GPT2        & 0.47 & 0.56 & 0.64 & 0.71 \\
                           & Mamba       & 0.48 & 0.57 & 0.64 & 0.71 \\     
                           & T5         & 0.49 & 0.57 & 0.65 & 0.72 \\ 
                            & T5\textsubscript{qrecc}         & 0.47 & 0.56 & 0.65 & 0.71 \\ 
                            & MiniLM        & \textbf{0.50} & \textbf{0.59} & \textbf{0.67} & \textbf{0.73} \\ 
                           \hline
\multirow{3}{1.5em}{PU}   
                           & GPT2        & 0.51 & 0.60 & 0.67 & 0.74 \\
                           & Mamba       & 0.51 & 0.60 & 0.67 & 0.74 \\  
                           & T5         & 0.52 & 0.62 & 0.69 & 0.76 \\
                           & T5\textsubscript{qrecc} & 0.52 & 0.61 & 0.69 & 0.75 \\
                            & MiniLM        & \textbf{0.54} & \textbf{0.63} & \textbf{0.71} & \textbf{0.78} \\ 

\end{tabular}
}
\caption{\textbf{Additional \rrf retrieval results} in \textbf{synthetic dialogs} generated via \textbf{propositions}, using proprietary (PR) and public (PU) documents, and leveraging the fine-tuned query rewriters and the fine-tuned retriever (\minilm). Results comparable to those of Table~\ref{tab:prop_vs_sent_private}.}
\vspace{-6mm}
\label{tab:main_IR}
\end{center}
\end{table}

\subsection{Retrieval in synthetic dialogs}
\label{sec:rewriters}
\label{main_experiments}
Next, following popular query rewriting approaches \cite{elgohary-etal-2019-unpack, lin-etal-2020-cqr, anantha-etal-2021-qrecc, li-gaussier-2024-domain}, we fine-tune three lightweight models to decontextualize user questions (\S\ref{sec:bulding}): 
\mamba 370\mparam \cite{gu-2024-mamba}, \gpttwo 350\mparam, \tfive 220\mparam \cite{raffel-etal-2020-t5}. No previous work explores Mamba for query rewriting; we use it because of its linear complexity and, thus, better performance on long sequences, compared to Transformers. We fine-tune the rewriters separately on the training sets of our synthetic dialogs. We show that these lightweight models, when fine-tuned, can achieve comparable results with prompting larger LLMs. Leveraging such models instead of larger LLMs, primarily during inference, can reduce the computational cost significantly. 
We also fine-tune the best performing model (\textsc{t5}) on a much larger/broader, manually generated dataset, \qrecc \cite{anantha-etal-2021-qrecc}, for comparison purposes. Finally, using our synthetic dialogs we fine-tune a lightweight Sentence Transformer (for the same reason as with the query rewriters), \minilm, as a retriever.\footnote{For \minilm the concatenation of the last user query and the dialog history is considered both during training and inference. Appendix \ref{sec:appendix_training} provides additional fine-tuning details.}

Table \ref{tab:main_IR} shows that lightweight rewriters perform similarly to each other and better than using contextualized user questions, with or without concatenating the dialog context (Context, Query\textsubscript{co} in Table~\ref{tab:prop_vs_sent_private}). Naturally, lightweight rewriters cannot outperform the `ground truth' decontextualized queries (Query\textsubscript{de}) of Dialog-\llm (Step 2). Additional results with \bm and dense retrieval are presented in Appendix~\ref{sec:appendix_bm25_dense}. Although smaller, \tfive has the best performance among the three lightweight rewriters; and \tfive trained on our synthetic dialogs slightly outperforms \tfive trained on \qrecc 
(\textsc{t5}\textsubscript{qrecc}), which was trained on almost 14 times more data.   
Based on its performance, we use \tfive as the only query rewriter in subsequent experiments. We also do not experiment further with Context, 
given its poor results (Tables \ref{tab:prop_vs_sent_private}--\ref{tab:main_IR}).

Table \ref{tab:main_IR} also shows that fine-tuning a retriever (\minilm) is more effective than query rewriting, matching the performance of `ground truth' decontextualized queries (Query\textsubscript{de} in Table \ref{tab:prop_vs_sent_private}). This contradicts previous reports that fine-tuning the retriever requires more data (\S\ref{sec:reformulation}), at least in our scenario. Hence, we continue to experiment with the fine-tuned retriever in the following experiments.

\subsection{Response generation in synthetic dialogs} \label{sec:response}

We now use the contextualized user questions (Query\textsubscript{co}), the decontextualized user questions of the \tfive rewriter, or the `ground-truth' decontextualized questions (Query\textsubscript{de}) as  queries to the \rrf retriever. We then feed the Response-\llm with the top-20 retrieved propositions and instruct it to generate the system response (prompt in Appendix~\ref{sec:appendix_a_step4}). We also retrieve propositions using our fine-tuned retriever (\minilm). Since the last user query is not self-contained we also provide Response-\llm with the dialog history, not just the last user query, in this case. Table~\ref{tab:main_RAG} compares the generated responses to the `ground-truth' system responses (generated by Dialog-\llm in Step 2).
The decontextualized questions of \tfive lead to better responses, compared to the contextualized ones. \minilm leads to equally good or better responses compared to Query\textsubscript{de}, despite their almost identical \textsc{ir} performance (Tables~\ref{tab:prop_vs_sent_private}--\ref{tab:main_IR}), presumably because the additional dialog history used with \minilm provides more useful information to Response-\llm than the `gold' decontextualized query of Query\textsubscript{de}.  

\begin{table}[b]
\begin{center}
\resizebox{\columnwidth}{!}{
\begin{tabular}{llllllll}
\hline
                           & \textbf{Method}                     & \textbf{SBLEU$\uparrow$}  & \textbf{METEOR$\uparrow$}  & \textbf{BSC$\uparrow$}  & \textbf{PL$\downarrow$}   \\ \hline
\multirow{3}{1.5em}{PR}   & Query\textsubscript{co} & 40.57 & 55.81   & 93.24 & 3.78 \\   
                           & T5          & 42.79 & 58.67   & 93.65 & 3.34 \\
                           & Query\textsubscript{de} & 44.52 & 59.72   & 93.99 & \textbf{3.25} \\ 
                           &MiniLM     & \textbf{46.14} & \textbf{60.11}           & \textbf{94.28} & 3.29 \\
                           \hline
\multirow{3}{1.5em}{PU}   & Query\textsubscript{co} & 44.92 & 58.99   & 93.39 & 3.03 \\  
                           & T5          & 47.33 & 62.11   & 93.80 & 2.74 \\
                           & Query\textsubscript{de} & 48.73 & \textbf{63.46}   & 94.08 & \textbf{2.68}\\
                           &MiniLM     & \textbf{49.63} & 62.55           & \textbf{94.33} & 2.70 \\
\end{tabular}
}
\caption{\textbf{Response generation results} in \textbf{synthetic dialogs} generated via \textbf{propositions} from proprietary (PR) and public (PU) documents, using \rrf retrieval.}
\label{tab:main_RAG}
\vspace{-5mm}
\end{center}
\end{table}

\subsection{Retrieval in real-world dialogs}

We now provide evaluation scores in the \emph{real-world} \doctodial and \multidoctodial datasets. We discard their training sets to demonstrate the value of our synthetic data in new application domains \emph{without any} manually annotated dialogs.

We use \tfive and \minilm fine-tuned on synthetic dialogs generated from the \doctodial documents (\S\ref{sec:rewriters}). Alternatively, we rewrite user questions by prompting an \llm (\claude, also used as Dialog-\llm). Note that ground-truth question rewrites are not available in \doctodial and \multidoctodial, 
but only ground-truth \emph{passages}, thus we also use \rrf for \emph{passage} retrieval.

Table~\ref{tab:doc2dial_IR} shows that \tfive substantially improves the retrieval performance in both datasets, compared to using contextualized questions (Query\textsubscript{co}). Prompting \textsc{claude} leads to further substantial improvements, at the cost of invoking an expensive \llm, while \minilm again matches its performance.  
For reference, Table~\ref{tab:doc2dial_IR} also includes the reported results of \citet{feng-etal-2020-doc2dial} and \citet{feng-etal-2021-multidoc2dial}, who directly \emph{fine-tune a dense retriever} on the training sets of the two datasets, thus \emph{requiring manually annotated domain-specific data}; hence, their results cannot be fairly compared to ours.

\begin{table}[t]
\begin{center}
\resizebox{\columnwidth}{!}{
\begin{tabular}{llllllll}
\hline
                \textbf{D2D/Approach} & \textbf{Method} & \textbf{MAP$\uparrow$}  & \textbf{R@5$\uparrow$}  & \textbf{R@10$\uparrow$} & \textbf{R@20$\uparrow$} \\ \hline
--                 & Query\textsubscript{co}  & 0.17 & 0.26           & 0.34          & 0.40 \\
Rewr-synFT    & T5                       & 0.21 & 0.31           & 0.41          & 0.48 \\
Rewr-prompt     & Claude  & \textbf{0.25} & \textbf{0.36}           & \textbf{0.47} & \textbf{0.56} \\
Retr-synFT     & MiniLM  & 0.24 & 0.35           & 0.46 & 0.55 \\
Retr-FT*   & \citet{feng-etal-2020-doc2dial}                  & n/a  & 0.85  & 0.90 & n/a   \\  
\hline
                            \textbf{MD2D/Approach} & \textbf{Method} & \textbf{MAP$\uparrow$}  & \textbf{R@5$\uparrow$}  & \textbf{R@10$\uparrow$} & \textbf{R@20$\uparrow$} \\ \hline
--   & Query\textsubscript{co} & 0.17 & 0.26 & 0.34 & 0.40 \\
Rewr-synFT  & T5 & 0.21 & 0.31 & 0.40 & 0.48 \\
Rewr-prompt     & Claude  & \textbf{0.23} & \textbf{0.34}           & \textbf{0.44} & \textbf{0.53} \\
Retr-synFT     & MiniLM  & \textbf{0.23} & \textbf{0.34}           & \textbf{0.44} & \textbf{0.53} \\
Retr-FT*   & \citet{feng-etal-2021-multidoc2dial}  &   n/a                   & n/a    & 0.69 & 0.79

\end{tabular}
}
\caption{\textbf{RRF passage retrieval results} in \textbf{real-world dialogs} from \doctodial (D2D) and \multidoctodial (MD2D).
T5/Claude: question rewritten by T5/Claude.
Rewr-synFT: rewriter fine-tuned on synthetic data, 
Rewr-prompt: the rewriter is a prompted \llm, 
Retr-synFT: retriever fine-tuned on synthetic data,
Retr-FT: retriever fine-tuned on \emph{manually annotated domain-specific data} (not comparable to our work). Starred results from \citet{feng-etal-2020-doc2dial} and \citet{feng-etal-2021-multidoc2dial}.
}
\vspace{-6mm}
\label{tab:doc2dial_IR}
\end{center}
\end{table}



\subsection{Response generation in real-world dialogs}
Table~\ref{tab:Multidoc2dial_RAG} shows response generation results using the real-world dialogs of the test set of \multidoctodial. We do not show results for \doctodial, as it concerns generating a system response from a single given document, which is incompatible with our synthetic data generation pipeline.
We now retrieve propositions, since entire documents or passages 
confuse the Response-\llm (\llamaeightb) with redundant information. Again, our \tfive rewriter and \minilm retriever (fine-tuned on synthetic data) improve performance, compared to using contextualized questions (Query\textsubscript{co}). Interestingly, decontextualizing questions by prompting \textsc{claude} 
does not necessarily lead to better response generation scores.
For completeness, we also include the method of \citet{feng-etal-2021-multidoc2dial}, who \emph{fine-tune} \textsc{bart} \cite{lewis-etal-2020-bart} for response generation using (their) \emph{manually annotated training data}; hence, their results are not directly comparable. 



\begin{table}[h!]
\begin{center}
\resizebox{\columnwidth}{!}{
\begin{tabular}{llllll}
\hline
                            \textbf{Approach} & \textbf{Method}                        & \textbf{SBLEU$\uparrow$}  & \textbf{METEOR$\uparrow$}  & \textbf{BSC$\uparrow$}  & \textbf{PL$\downarrow$}   \\ \hline
-- & Query\textsubscript{co}  & 6.16  & 20.93   & 85.63 & 26.04 \\
 Rewr-synFT & T5                       & 6.54  & 22.65   & 85.74 & 23.08 \\
 Rewr-prompt & Claude & 6.52 & 23.34 & 85.47 & 21.75 \\
 Retr-synFT     & MiniLM  & \textbf{8.89} & \textbf{26.66}           & \textbf{86.18} & \textbf{16.28} \\
 Retr-FT$^*$   & \citet{feng-etal-2021-multidoc2dial} & 21.9 & n/a & n/a & n/a \\

\end{tabular}
}
\caption{\textbf{Response generation results} in \textbf{real-world dialogs} from \multidoctodial, using \rrf retrieval of \textbf{propositions}. Responses generated by \llamaeightb in our (the first four) methods. Starred results from \citet{feng-etal-2021-multidoc2dial}. We use the same notation as in Table~\ref{tab:doc2dial_IR}.}
\vspace{-5mm}
\label{tab:Multidoc2dial_RAG}
\end{center}
\end{table}

\subsection{Conditional question rewriting}

Finally, we propose a new joint question classification/rewriting approach to reduce the expected latency in real-world applications. 
Again, we use the \tfive question rewriter (\S\ref{sec:rewriters}), but now during training we
prepend each (gold output) decontextualized question with the tokens `rewrite', if it is different from the contextualized one, or `no\_rewrite' otherwise. 
At inference, we stop the generation procedure if the `no\_rewrite' token is generated, and replace the token with the input (contextualized) question as the prediction of the question rewriter. We find no difference in performance between the new approach and \tfive of Table~\ref{tab:main_IR}. However, the average generation time for proprietary dialogs is reduced from 0.19 seconds to 0.09 
and for public dialogs from 0.24 seconds to 0.1. 
The reader is reminded (\S\ref{prop_vs_sent}) that 36\% of synthetic proprietary and 27\% public user questions require rewriting. 

\subsection{Human Evaluation Remarks for the Generated Dialogs}
Although our pipeline constitutes a simple concept (including two steps only), it involves prompting an LLM (\dialogllm) multiple times, making human evaluation of the LLM's responses challenging (hence the automatic evaluation presented in Table \ref{tab:quality}). In terms of proposition generation (Step 1), manual evaluation on a small sample (around 15\%) of the generated data revealed little to no hallucinated information, while the propositions themselves reflect the document knowledge to a surprisingly satisfactory level, in accordance with the findings of \citet{chen-etal-2024-dense}. However, cases of propositions that convey no information (e.g., expressions like ``Contact us.'') or information reflecting knowledge not exclusive to the documents themselves (e.g., where the settings of a particular browser are located) were also found. Finally, the generated propositions tend to be concise and exhibit little lexical diversity compared to the corresponding original parts of the documents. In terms of dialog generation (Step 2), no hallucinations were found, mostly owing to the filtering mechanism and the proposition substitution approach explained in Prompt 2.3 of Section \ref{proposed_pipeline}. An evaluation of the generated dialogs was also presented in Subsection \ref{prop_vs_sent}.


\section{Conclusions and Future Work}
We presented a new pipeline that generates annotated document-grounded dialogs, to alleviate the lack of training data in new application domains, requiring only a set of relevant domain-specific documents.
We highlighted the importance of using propositions, rather than  document sentences, for dialog generation, and showed 
that they lead to synthetic dialogs that are clearly superior in our experimental evaluation.
Using only our synthetic data, we trained both light question rewriters and a retriever. We showed that they substantially improve performance, compared to using the original questions with or without dialog history, and that their performance is comparable to obtaining rewrites by prompting \llm{s}. 
We also introduced a joint efficient question classifier/rewriter. 
In future work, we plan to explore means of generating more dialogs given the same set of documents (e.g., data augmentation), which will allow us to fine-tune the response-generator as well.
Finally, as dialogs in low resource languages are scarcer still, we plan to extend our pipeline to such languages, exploiting multilingual \llm{s} or machine translation.

\section{Limitations}
\label{sec:limitations}
A limitation of our work is its dependence on large costly \llm{s} (like \claude) for the creation of synthetic data. However costly these models may be, we deem their usage essential to ensure high quality synthetic data, as in previous work \cite{chen-etal-2024-sub, chen-etal-2024-dense, mo-etal-2024-convsdg}. Applying our pipeline to large volumes of data may pose a problem cost-wise; the application may even be infeasible for datasets containing millions of passages, as in the case of the \qrecc dataset \cite{anantha-etal-2021-qrecc}. Nevertheless, in many realistic scenarios, like our proprietary data or the \doctodial/\multidoctodial datasets, the cost of executing our pipeline is negligible.

Our generated data, being purely synthetic, 
may contain errors in the form of hallucinations. For example, there is no guarantee that the generated propositions (Step 1 of the generation pipeline) perfectly reflect the knowledge in the documents; in our experiments, however, such cases are scarce. 


Finally, although we showed how the generated synthetic data can be used to train lightweight question rewriters or retrievers, 
response generation still relies on prompting \llm{s}. Preliminary experiments (not reported) showed that light response generators (e.g., \tfive) fine-tuned on our current synthetic data severely under-perform compared to prompting \llm{s} as response generators, possibly because the synthetic datasets are not large enough. We, hence, left this direction for future work.  

\section{Ethical Considerations}
\label{sec:ethical}
A major concern regarding \llm{s} like \claude, which our generation pipeline leverages, is that sensitive data may be stored by third parties and may even be exposed publicly. In our case, we either used already publicly available documents, or documents that do not include sensitive information and their processing has been approved by qualified individuals. We advise potential users of the pipeline to take similar precautions. 

\section{Acknowledgments}
\label{sec:acknowledgements}
This work was partially supported by: Omilia - Conversational Intelligence; the Research Center of the Athens University of Economics and Business (AUEB-RC); and project MIS 5154714 of the National Recovery and Resilience Plan Greece 2.0 funded by the European Union under the NextGenerationEU Program. This work has received funding from the European Union’s Horizon Europe research and innovation programme under the project ELOQUENCE (Grant Agreement No. 101135916). Views and opinions expressed are however those of the author(s) only and do not necessarily reflect those of the European Union or European Commission-EU. Neither the European Union nor the granting authority can be held responsible for them.\\
The authors thank Panagiotis Tassias, Nikolaos Kokkinis-Ntrenis and Katerina Tzortzi from Omilia for sharing their experience in dataset creation. 
\begin{figure}[H]
\begin{center}
\includegraphics[scale=0.22]{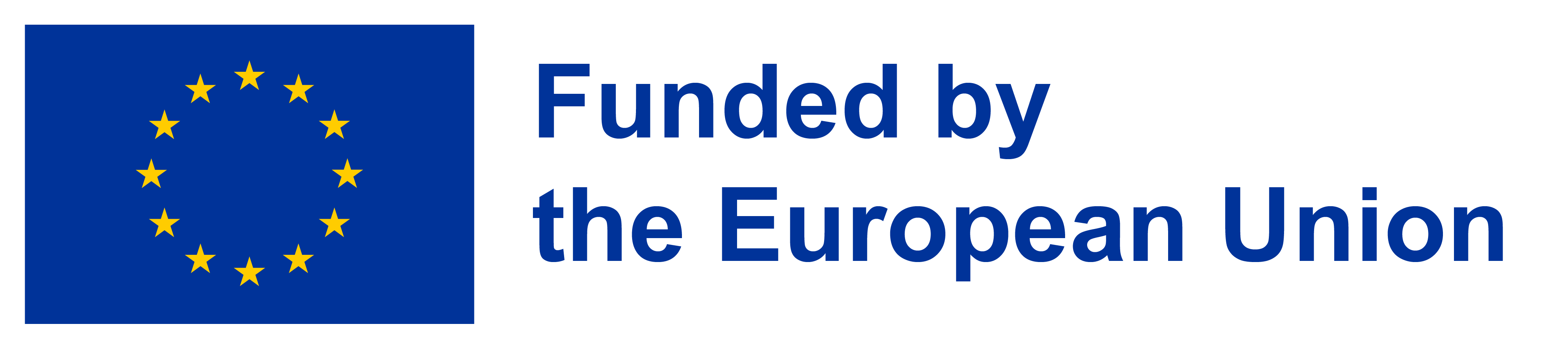} 
\vspace{-3mm}
\label{figures/EN_FundedbytheEU_RGB_POS.png}
\vspace{-5mm}
\end{center}
\end{figure}

\bibliography{acl_latex}

\appendix
\section*{Appendix}
\section{Examples of Synthetic Dialogs}
\label{sec:appendix_dial}

Below are snippets of two synthetic dialogs generated from the \doctodial data. Each dialogic pair contains a contextualized and decontextualized user turn (\user, \user decont.) and the system response (\system). Notice that user turns 2--4 and turn 6 of Dialog 7 make references to user turn 1 (`it' refers to `Board Appeal', as can be seen in the corresponding \user decont.\ turns). Moreover, Turn 8 mentions a (`Board Appeal') decision, omitting `Board Appeal', which is mentioned in previous turns. Similarly, for Dialog 356, Turns 2--4 make references to `HazMat endorsement', without explicitly mentioning it. Notice that users may ask multiple questions in their turns (Turn 3).\\

\noindent\texttt{---- Synthetic Doc2Dial Dialog id 7 ---- \\ \\ }
\noindent\footnotesize\texttt{---- 1 ---- \\
\textbf{USER         :} How can I submit a Decision Review Request for a Board Appeal? \\
\textbf{USER decont. :} How can I submit a Decision Review Request for a Board Appeal? \\
\textbf{SYSTEM       :} To submit the Decision Review Request: Board Appeal VA Form 10182, you can apply by mail, in person, or by fax. \\
}

\noindent\footnotesize\texttt{---- 2 ---- \\
\textbf{USER         :} What are the steps to apply for it by mail? \\
\textbf{USER decont. :} What are the steps to apply for a Board Appeal by mail? \\
\textbf{SYSTEM       :} To apply for a Board Appeal by mail, you need to send the completed VA Form 10182 to the address: Board of Veterans Appeals, PO Box 27063, Washington, D.C. 20038. \\
}

\noindent\footnotesize\texttt{---- 3 ---- \\
\textbf{USER         :} How can I apply for it in person? \\
\textbf{USER decont. :} How can I apply for a Board Appeal in person? \\
\textbf{SYSTEM       :} To apply for a Board Appeal in person, you need to bring your completed VA Form 10182 to a regional benefit office. \\
}

\noindent\footnotesize\texttt{---- 4 ---- \\
\textbf{USER         :} Can I apply for it by fax? \\
\textbf{USER decont. :} Can I apply for a Board Appeal by fax? \\
\textbf{SYSTEM       :} Yes, to apply for a Board Appeal by fax, you need to fax your completed VA Form 10182 to 844-678-8979. \\
}

\noindent\footnotesize\texttt{---- 5 ---- \\
\textbf{USER         :} How can I get a copy of VA Form 10182 to fill out for a Board Appeal? \\
\textbf{USER decont. :} How can I get a copy of VA Form 10182 to fill out for a Board Appeal? \\
\textbf{SYSTEM       :} You can ask a regional benefit office for a copy of VA Form 10182 to fill out for a Board Appeal. You can also call the VA toll-free hotline at 800-827-1000, Monday through Friday, 8:00 a.m. - 9:00 p.m. ET, to request VA Form 10182 for a Board Appeal. \\
}

\noindent\footnotesize\texttt{---- 6 ---- \\
\textbf{USER         :} What is the time limit to request it? \\
\textbf{USER decont. :} What is the time limit to request a Board Appeal? \\
\textbf{SYSTEM       :} You have 1 year from the date on your decision to request a Board Appeal. However, if you're one of multiple people claiming the same benefit (which is rare), you have 60 days to request a Board Appeal. \\
}

\noindent\footnotesize\texttt{---- 7 ---- \\
\textbf{USER         :} What can I do if I disagree with the Board's decision? \\
\textbf{USER decont. :} What can I do if I disagree with the Board's decision? \\
\textbf{SYSTEM       :} If you disagree with the Board's decision and have new and relevant evidence that supports your case, you can file a Supplemental Claim after a Board decision. You can also appeal to the U.S. Court of Appeals for Veterans Claims after a Board decision. \\
}

\noindent\footnotesize\texttt{---- 8 ---- \\
\textbf{USER         :} Do I need to do anything while waiting for the decision? \\
\textbf{USER decont. :} Do I need to do anything while waiting for the Board Appeal decision? \\
\textbf{SYSTEM       :} After you request a decision review like a Board Appeal, you don't need to do anything while you wait unless VA sends you a letter asking for more information. However, if VA schedules exams for you after you request a decision review like a Board Appeal, be sure not to miss them. \\ \\
}

\noindent\normalsize\texttt{---- Synthetic Doc2Dial Dialog id 356 ---- \\  }

\noindent\footnotesize\texttt{---- 1 ---- \\
\textbf{USER         :} What is the validity period of a CDL with HazMat endorsement issued by the DMV if the background checks are not completed before the endorsement expires? \\
\textbf{USER decont. :} What is the validity period of a CDL with HazMat endorsement issued by the DMV if the background checks are not completed before the endorsement expires? \\
\textbf{SYSTEM       :} The DMV will provide a CDL with the HazMat endorsement that is valid for 90 days only if you applied to renew a current NY State CDL with a HazMat endorsement and the background checks are not completed and approved before the endorsement expires. \\
}

\noindent\footnotesize\texttt{---- 2 ---- \\
\textbf{USER         :} What happens if I don't pass the written test or background checks for it? \\
\textbf{USER decont. :} What happens if I don't pass the written test or background checks for the HazMat endorsement? \\
\textbf{SYSTEM       :} If you do not pass the written test or the background checks for the HazMat endorsement, the DMV will issue a CDL that does not have a HazMat endorsement. \\
}

\noindent\footnotesize\texttt{---- 3 ---- \\
\textbf{USER         :} How will I be notified if I am approved for it? And what should I do after receiving the notification? \\
\textbf{USER decont. :} How will I be notified if I am approved for the HazMat endorsement? And what should I do after receiving the notification? \\
\textbf{SYSTEM       :} The DMV will mail you notification of approval for the HazMat endorsement along with instructions to go into a DMV office to obtain a CDL with the HazMat endorsement. \\
}

\noindent\footnotesize\texttt{---- 4 ---- \\
\textbf{USER         :} What information will be displayed on my new CDL with it? \\
\textbf{USER decont. :} What information will be displayed on my new CDL with the HazMat endorsement? \\
\textbf{SYSTEM       :} Your new CDL with the HazMat endorsement will display the expiration date of your CDL and the expiration date of your HazMat endorsement. \\
}

\normalsize
\section{Prompts Used to Generate Dialogs}
\label{sec:appendix_a}

\subsection{Step 1 prompt}
\label{sec:appendix_a_step1}
\footnotesize
\hfill \break
\noindent\texttt{Read the document you will be given 
and look for questions and answers 
in it. Return propositions 
if the document includes information 
that could actually answer user 
questions. If the document only 
has links or vague information that 
can't answer questions, do not return 
propositions. Also, do not return 
propositions if the document only has 
questions. If the document does have 
questions and answers, break them down 
into simple and clear propositions 
that make sense on their own. 
Recognize the language of the document 
given below and provide the 
propositions in the original language 
as the given Document.}

\noindent\texttt{If you do not create propositions the 
reply must be an empty list such as [] and 
nothing else.}

\begin{verbatim}
Here is a document:
<document>
{text}
</document>
\end{verbatim}

\noindent\texttt{To generate propositions you need to:}

\noindent\texttt{1. Split compound sentence into simple 
English sentences. Maintain the original 
phrasing from the input whenever 
possible.}

\noindent \texttt{2. For any named entity that is 
accompanied by additional descriptive 
information, separate this information 
into its own distinct proposition.}

\noindent \texttt{3. Decontextualize the proposition by 
adding necessary modifier to nouns or 
entire sentences and replacing pronouns 
(e.g., "it", "he", "she", "they", 
"this", "that") with the full name 
of the entities they refer to.}

\noindent \texttt{4. Present the results as a list of 
strings, formatted in JSON. Provide 
only the JSON and nothing else.}

\normalsize
\subsection{Step 2 prompts}
\label{sec:appendix_a_step2}

\textbf{Prompt 2.1}


\noindent To maintain the dialog flow, we instruct the model to keep relevant (to each other) queries in adjacent turns. We also encourage the \llm to generate queries that are grounded in more than one proposition. The purpose of the first two instructions, which are related to turns where the user and system exchange greetings, is to mimic real dialogs, but can be skipped without affecting the quality of the dialogs. The output is a \textsc{json} dictionary of question-answer pairs, each containing the decontextualized query and its answer.     

\footnotesize
\hfill \break
\noindent\texttt{Your task is to read the given propositions 
and generate a dialog between a user and 
a system, where the user asks certain 
questions and the system tries to provide 
answers.} 

\noindent\texttt{Follow these instructions:}

\noindent\texttt{1. Your response should be a JSON of 
the following format:}

\begin{verbatim}  
{
  "0" : {
    "<user>": ,
    "<system>":,
  },
  "1" : {
    "<user>": ,
    "<system>":,
  },
  ...
}
\end{verbatim}

\noindent\texttt{2. The dialog must start with the 
user greeting the system and the 
system replying politely.}

\noindent\texttt{3. The dialog must end with user 
thanking the system and the system 
replying politely.}

\noindent\texttt{4. In each dialog turn, the user 
asks a question based on a given 
proposition. The user question must 
be a self-contained, standalone 
question without the need to refer 
to previous dialog context.}

\noindent\texttt{5. A user may also ask complex 
questions, for which the answer 
can be two or more propositions.}

\noindent\texttt{6. In each dialog exchange the 
system answers the user question 
based on the propositions.}

\noindent\texttt{7. Make sure that the user 
questions referring to the same 
propositions are in adjacent 
turns.}

\noindent\texttt{8. Each system's answer must be 
a full sentence.}

\begin{verbatim}
<propositions>
{}
</propositions>
\end{verbatim}

\normalsize
\hfill \break
\noindent\textbf{Prompt 2.2}


\footnotesize
\hfill \break
\noindent\texttt{Your task is to read the given dialog. 
The dialog you will be given has a JSON 
format. The key <user> refers to user 
utterances, while the key <system> refers 
to the system utterances.
Make the user utterances dependent on 
previous dialog turns taking into 
account the dialog context and using 
pronouns to replace already mentioned 
information only if such information 
is already mentioned in the previous 
dialog turns.
Only return a JSON of the following 
format:}

\begin{verbatim}
{
  "0" : {
    "<contextualized user>": ,
    "<system>":
  },
  "1" : {
    "<contextualized user>": ,
    "<system>":
  },
  ...
}

Here is the dialog:
<dialog>
{}
</dialog>
\end{verbatim}

\normalsize
\hfill \break
\noindent\textbf{Prompt 2.3}
\hfill \break
\footnotesize
\noindent\texttt{I will give you a list of propositions 
and a text in JSON format of 
question and answer pairs generated 
from these propositions.
I need you to act as a human annotator 
and evaluate the question and answer 
pairs provided following these 
instructions:}

\noindent\texttt{1. Provide a separate review and evaluation 
for each question and answer.}

\noindent\texttt{2. First check if the questions provided 
are correctly generated from the 
propositions provided.}

\noindent\texttt{3. The answer to each question should 
be reflecting the information provided 
in the propositions.}

\noindent\texttt{4. Note which propositions are used 
in each answer.}

\noindent\texttt{5. If a question and answer is generated 
from the provided propositions after 
your review, mark it as "accepted". 
If not, mark it as "not\_accepted".}

\noindent\texttt{6. The first and last pairs should 
always be accepted.}

\noindent\texttt{7. Return only a dictionary in JSON 
format and nothing else. 
The key of each dictionary should be 
the same with each question answer 
pair given. Follow the example:}

\begin{verbatim}
{
  "0": {
    "propositions_used":  
    ,
    "explain_evaluation": ,
    "evaluation": ,

  },
}
\end{verbatim}   
\noindent\texttt{Here are the propositions 
and the question-answer pairs:}

\begin{verbatim}    
<propositions>
{}
</propositions>

<question and answer pairs>
{}
</question and answer pairs>
\end{verbatim}

\normalsize
\subsection{Response-\llm prompt}
\label{sec:appendix_a_step4}

Having rewritten the user question and having retrieved relevant propositions, we prompt \llamaeightb instruct to generate a system response, conditioned on the top-20 retrieved propositions and the rewritten query. If the query cannot be answered using the provided propositions, the \llm is instructed to generate the token <cannot\_answer>. 
\hfill \break
Alternatively, when we use the fine-tuned (dialog history-aware) retriever (\minilm), we provide the whole dialog history to Response-\llm, along with the last user question and inform the model that it should also consider the additional context. 

\medskip
\footnotesize
\noindent\texttt{Your job is to answer user questions given a set of propositions in a 
list format. There may be irrelevant propositions included.} 

\noindent\texttt{You only need to provided the answer.
If the question cannot be answered 
using the provided propositions, 
generate the token <cannot\_answer> only.} 

\begin{verbatim}    
Here are the propositions: {}
\end{verbatim}

\begin{verbatim}
Here is the user question: {} 
\end{verbatim}

\normalsize

\section{Statistics of Synthetic Dialogs}
\label{sec:appendix_stats}
Additional statistics for the 614 proprietary and 420 public dialogs generated through our pipeline are presented in Table \ref{tab:add_stats}.

\begin{table}[b]
\begin{center}
\resizebox{\columnwidth}{!}{
\begin{tabular}{llll}
\hline
                           & \textbf{Query}        & \textbf{Mean}  & \textbf{STD}   \\ \hline
\multirow{4}{1.5em}{PR} & Document length &316.35&243.95\\
                        & Proposition length & 19.21 & 26.97\\
                        & QA pairs per dialog &10.68 &3.85\\
                        & GT Propositions per QA pair & 2.04 & 1.54\\
                        \hline   
\multirow{4}{1.5em}{PU} & Document length &798.32&508.72\\
                        & Proposition length & 23.30 & 13.95\\
                        & QA pairs per dialog &14.58 &5.85\\
                        & GT Propositions per QA pair & 1.99& 1.55\\
                        \hline   
\end{tabular}
}
\caption{\textbf{Additional statistics} (mean and standard deviation) \textbf{of the generated dialogs} using proprietary (PR) and public (PU) documents, including the length (in number of words) of the documents, generated propositions, dialogs, and the number of ground truth (GT) propositions per QA pair.}
\label{tab:add_stats}
\end{center}
\end{table}

\section{Training Details}
\label{sec:appendix_training}
For every experiment in which a model requires training (fine-tuning), we reserve roughly 25\% of the training data as our validation set.
Our query rewriters (\mamba, \gpttwo, \tfive), are trained with a batch size of 8 training instances, for 100 epochs or until no improvement is observed on the validation set for 15 consecutive steps, with a learning rate of $10^{-4}$. Moreover, if no improvement is observed after 10 consecutive steps, we reduce the learning rate to $10^{-5}$. Similarly, our retriever is trained using the same early stopping approach, with a batch size of 16 and $10^{-5}$ learning rate. After 10 steps, if no improvement is observed in the validation set, the learning rate is reduced to $10^{-6}$.

\section{Additional Retrieval Results}
\label{sec:appendix_bm25_dense}

We present results of \bm and dense retrieval for proprietary and \doctodial documents, separately, in Table \ref{tab:main_multiple_retrievers_private}.
Both retrievers exhibit the same behavior as the \rrf retriever (Tables~\ref{tab:prop_vs_sent_private}--\ref{tab:main_IR}); \tfive manages to outperform the contextualized query, but not the `ground-truth' decontextualized one. Both retrievers perform worse than the \rrf retriever.

\begin{table}[h!]
\begin{center}
\resizebox{\columnwidth}{!}{
\begin{tabular}{llllllll}
\hline
                          \textbf{PR} &                      & \textbf{MAP$\uparrow$}  & \textbf{R@5$\uparrow$}  & \textbf{R@10$\uparrow$} & \textbf{R@20$\uparrow$}\\ \hline
\multirow{3}{3em}{Dense} & Query\textsubscript{co} & 0.41 & 0.51 & 0.57 & 0.63 \\
                           & T5          & 0.46 & 0.54 & 0.61 & 0.66 \\
                           & Query\textsubscript{de} & \textbf{0.48} & \textbf{0.58} & \textbf{0.65} & \textbf{0.72} \\ \hline
\multirow{3}{3em}{\bm}   & Query\textsubscript{co} & 0.42 & 0.51 & 0.57 & 0.63 \\
                           & T5          & 0.45 & 0.54 & 0.61 & 0.66 \\
                           & Query\textsubscript{de} & 0.47 & 0.56 & 0.63 & 0.68 \\
\hline
                           \textbf{PU} &                     & \textbf{MAP$\uparrow$}  & \textbf{R@5$\uparrow$}  & \textbf{R@10$\uparrow$} & \textbf{R@20$\uparrow$}\\ \hline

\multirow{3}{3em}{Dense} & Query\textsubscript{co} & 0.45 & 0.54 & 0.62 & 0.69 \\
                           & T5          & 0.49 & 0.57 & 0.67 & 0.73 \\
                           & Query\textsubscript{de} & \textbf{0.51} & \textbf{0.60} & \textbf{0.68} & \textbf{0.74} \\ \hline
\multirow{3}{3em}{\bm}   & Query\textsubscript{co} & 0.45 & 0.53 & 0.60 & 0.66 \\
                           & T5          & 0.48 & 0.56 & 0.63 & 0.70 \\
                           & Query\textsubscript{de} & 0.49 & 0.59 & 0.65 & 0.71\\
\end{tabular}
}
\caption{\textbf{Dense and \bm retrieval results} in \textbf{synthetic dialogs} generated via \textbf{propositions}, using proprietary
(PR) and public (PU) documents.}
\label{tab:main_multiple_retrievers_private}
\end{center}
\end{table}

\section{Experiments in Separate Domains}
\label{sec:appendix_b}
In many real-life scenarios, the domain of each question is known during inference; the user may also explicitly request documents for a particular domain. Thus, the retriever only needs to consider information of the corresponding domain. To simulate such a scenario, we split propositions and questions based on their domains. Both proprietary and \doctodial datasets include documents and questions from four distinct domains (\S\ref{sec:documents}). The results are presented in Tables~\ref{tab:performance_domain_private} and \ref{tab:performance_domain_public} for proprietary and \doctodial dialogs, respectively. Regardless of dataset or domain, we reach similar conclusions regarding the performance of the question rewriter as in the main text (\S\ref{main_experiments}); the question rewriter outperforms using the original contextualized queries, but not the `ground-truth'  decontextualized queries. The fine-tuned retriever retains the best performance. 

In the proprietary data, we notice a drop in performance for the Miscellaneous (Misc) and Finance domains, compared to the main experiments (Table~\ref{tab:main_IR}). This is mostly due to the more complex and much longer user questions of these two domains. A similar observation can be made for the \ssa domain of \doctodial. For the rest of the domains of both datasets, the performance is equal to, or better than the performance reported in the main experiments, which is to be expected, since the retriever has to consider fewer propositions.

\begin{table}[h]
    \centering
    \resizebox{\columnwidth}{!}{
    \begin{tabular}{|lllll|}
        \hline
        & \textbf{MAP$\uparrow$} &  \textbf{R@5$\uparrow$} & \textbf{R@10$\uparrow$} & \textbf{R@20$\uparrow$}\\
        \hline
        \multicolumn{5}{|c|}{\textbf{Finance}} \\
        \hline
        Query\textsubscript{co} & 0.38 & 0.44 & 0.51 & 0.57 \\
        T5          & 0.39 & 0.46 & 0.52 & 0.59 \\
        Query\textsubscript{de} & 0.40 & 0.48 & 0.54 & 0.60 \\
        MiniLM          & \textbf{0.41} & \textbf{0.49} & \textbf{0.56} & \textbf{0.61} \\
        \hline
        \multicolumn{5}{|c|}{\textbf{Software}} \\
        \hline
        Query\textsubscript{co} & 0.45  & 0.56 & 0.63 & 0.71 \\
        T5          & 0.49  & 0.58 & 0.65 & 0.73 \\
        Query\textsubscript{de} & \textbf{0.50}  & 0.60 & 0.67 & 0.74\\
        MiniLM          & \textbf{0.50} & \textbf{0.62} & \textbf{0.71} & \textbf{0.78} \\
        \hline
        \multicolumn{5}{|c|}{\textbf{Insurance}} \\
        \hline
        Query\textsubscript{co} & 0.59 & 0.70 & 0.75 & 0.79 \\
        T5          & 0.59 & 0.71 & 0.76 & 0.79 \\
        Query\textsubscript{de} & \textbf{0.62} & \textbf{0.72} & \textbf{0.77} & 0.79 \\
        MiniLM          & \textbf{0.62} & \textbf{0.72} & \textbf{0.77} & \textbf{0.81} \\
        \hline
        \multicolumn{5}{|c|}{\textbf{Misc}} \\
        \hline
        Query\textsubscript{co} & 0.30 & 0.36 & 0.44 & 0.49 \\
        T5          & 0.30 & 0.36 & 0.45 & 0.46 \\
        Query\textsubscript{de} & 0.33 & \textbf{0.41} & \textbf{0.49} & \textbf{0.50} \\
        MiniLM          & \textbf{0.40} & \textbf{0.41} & 0.44 & 0.46 \\
        \hline
    \end{tabular}
    }
    \caption{\textbf{Retrieval results} in \textbf{synthetic dialogs} generated via \textbf{propositions} from \textbf{proprietary documents}, \textbf{separately for each domain}.}
    \label{tab:performance_domain_private}
\end{table}

\begin{table}[h]
    \centering
    \resizebox{\columnwidth}{!}{
    \begin{tabular}{|lllll|}
        \hline
        & \textbf{MAP$\uparrow$} &  \textbf{R@5$\uparrow$} & \textbf{R@10$\uparrow$} & \textbf{R@20$\uparrow$}\\
        \hline
        \multicolumn{5}{|c|}{\textbf{DMV}} \\
        \hline
        Query\textsubscript{co} & 0.56 & 0.65 & 0.73 & 0.78 \\
        T5          & 0.59 & 0.67 & 0.74 & 0.81 \\
        Query\textsubscript{de} & 0.60 & 0.68 & 0.75 & 0.81 \\
        MiniLM          & \textbf{0.63} & \textbf{0.72} & \textbf{0.80} & \textbf{0.84} \\
        \hline
        \multicolumn{5}{|c|}{\textbf{VA}} \\
        \hline
        Query\textsubscript{co} & 0.45  & 0.54 & 0.62 & 0.69 \\
        T5          & 0.50  & 0.59 & 0.68 & 0.76 \\
        Query\textsubscript{de} & 0.52  & 0.61 & 0.71 & 0.78\\
        MiniLM          & \textbf{0.57} & \textbf{0.66} & \textbf{0.74} & \textbf{0.83} \\
        \hline
        \multicolumn{5}{|c|}{\textbf{SSA}} \\
        \hline
        Query\textsubscript{co} & 0.44 & 0.56 & 0.63 & 0.69 \\
        T5          & 0.45 & 0.56 & 0.63 & 0.70 \\
        Query\textsubscript{de} & 0.46 & 0.57 & 0.68 & 0.71 \\
        MiniLM          & \textbf{0.49} & \textbf{0.61} & \textbf{0.69} & \textbf{0.75} \\
        \hline
        \multicolumn{5}{|c|}{\textbf{StudentAid}} \\
        \hline
        Query\textsubscript{co} & 0.54 & 0.62 & 0.70 & 0.77 \\
        T5          & 0.56 & 0.64 & 0.72 & 0.79 \\
        Query\textsubscript{de} & 0.57 &0.67 & 0.73 & 0.79 \\
        MiniLM          & \textbf{0.63} & \textbf{0.70} & \textbf{0.77} & \textbf{0.85} \\
        \hline
    \end{tabular}
    }
    \caption{\textbf{\rrf retrieval results} in \textbf{synthetic dialogs} generated via \textbf{propositions} from \textbf{public documents} (\doctodial), \textbf{separately for each domain}.}
    \label{tab:performance_domain_public}
\end{table}

\end{document}